\documentclass[10pt,twocolumn,letterpaper,dvipsnames,table]{article}
\usepackage[pagenumbers]{wacv} % To force page numbers, e.g. for an arXiv version

\usepackage[utf8]{inputenc} % allow utf-8 input
\usepackage[T1]{fontenc}    % use 8-bit T1 fonts
\usepackage{url}            % simple URL typesetting
\usepackage{booktabs}       % professional-quality tables
\usepackage{amsfonts}       % blackboard math symbols
\usepackage{nicefrac}       % compact symbols for 1/2, etc.
\usepackage{microtype}      % microtypography
\usepackage{xcolor}        % colors
\usepackage{amssymb}% http://ctan.org/pkg/amssymb
\usepackage{pifont}% http://ctan.org/pkg/pifont
\newcommand{\cmark}{\textcolor{Green}{\ding{51}}}%
\newcommand{\xmark}{\textcolor{red}{\ding{55}}}%
\usepackage[toc,page]{appendix}
\usepackage{subcaption}
\usepackage[font=small,labelfont=bf]{caption}
\usepackage{listings}
\usepackage{fancyvrb}

\definecolor{wacvblue}{rgb}{0.21,0.49,0.74}
\usepackage[pagebackref,breaklinks,colorlinks,allcolors=wacvblue]{hyperref}

\def\wacvPaperID{462} % *** Enter the WACV Paper ID here
\def\confName{WACV}
\def\confYear{2026}

\title{\underline{GeneVA}: A Dataset of Human Annotations for \underline{Gene}rative Text to \underline{V}ideo \underline{A}rtifacts}

\author{
    Jenna Kang \quad
    Maria Beatriz Silva \quad
    Patsorn Sangkloy \quad
    Kenneth Chen \quad
    Niall L. Williams \quad
    Qi Sun \\
    New York University \\
    {\tt\small \{jennakang, ms14127, ps5688, kennychen, n.williams, qisun\}@nyu.edu}
}

\begin{document}
\maketitle

\begin{abstract}
Recent advances in probabilistic generative models have extended capabilities from static image synthesis to text-driven video generation. However, the inherent randomness of their generation process can lead to unpredictable artifacts, such as impossible physics and temporal inconsistency. 
Progress in addressing these challenges requires systematic benchmarks, yet existing datasets primarily focus on generative images due to the unique spatio-temporal complexities of videos.
To bridge this gap, we introduce GeneVA, a large-scale artifact dataset with rich human annotations that focuses on spatio-temporal artifacts in videos generated from natural text prompts. We hope GeneVA can enable and assist critical applications, such as benchmarking model performance and improving generative video quality.
\end{abstract}
    
% \section{Introduction}

% Test citation: \cite{wang2024vidprom}

% Test citation: \citep{wang2024vidprom}

% Test citation: \citet{wang2024vidprom}

% \cite{wang2024vidprom}

\section{Introduction}
\label{sec:introduction}

%\input{figs/teaser}

% \note{
% What problem we are trying to solve.
% Why it is important, and why people should care.
% }%note
Recent advancements in diffusion models have significantly advanced text-driven visual generation, extending capabilities from static image generation to dynamic video content \cite{ho2022video,xing2024survey,guoanimatediff,khachatryan2023text2video}. 
Despite these breakthroughs, state-of-the-art text-video generative models often introduce a number of unique artifacts no present in still images \cite{chen2023watching}.
In AI-generated videos, both spatial (e.g., distorted geometry or inconsistent appearance) and temporal (e.g., physically implausible motion or temporal incoherence) artifacts arise.
A fundamental roadblock in characterizing and mitigating these unique artifacts is the lack of data and metrics to assess them.
Crucially, while metrics \cite{heusel2017gans,zhang2018unreasonable} may capture some sense of overall quality, assessment by real human observers is the highest standard for quality evaluation.

\begin{figure}[t]
\centering
\includegraphics[width=\linewidth,trim={.8cm 7.7cm .8cm 7cm},clip]{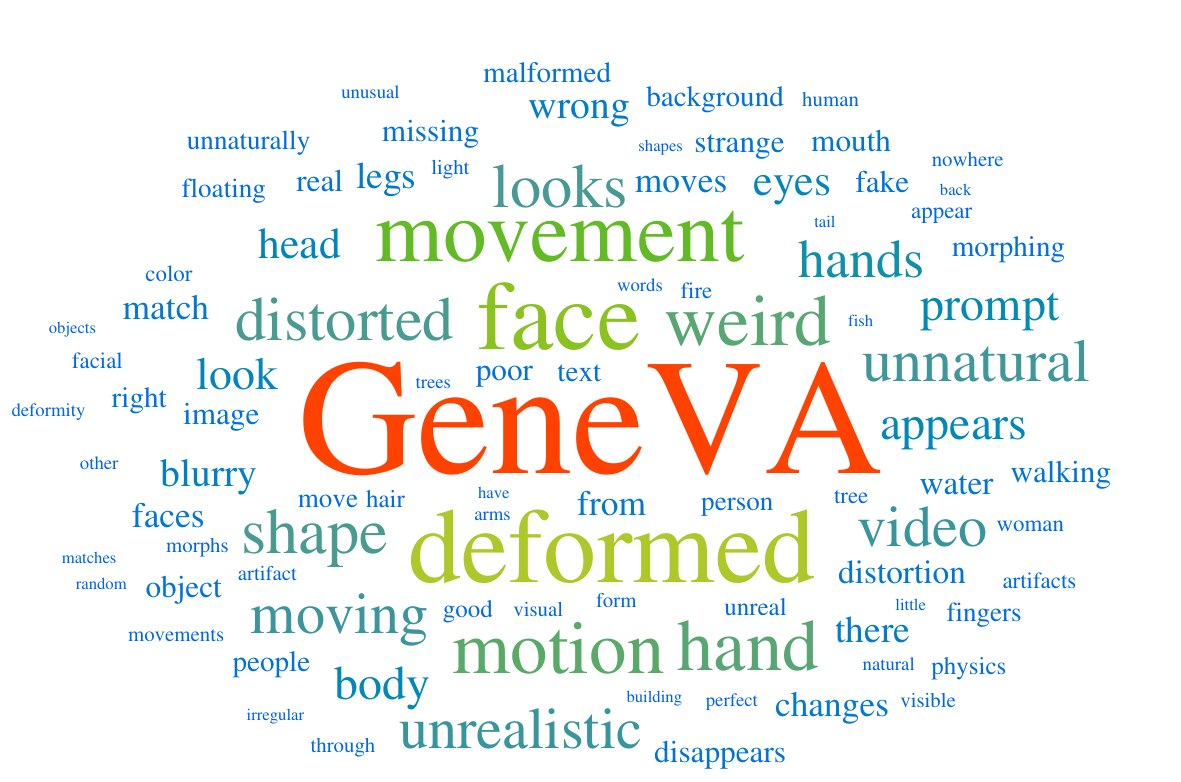}
\caption{
Human annotators were asked to describe, in a free text form, artifacts in AI-generated video.
A word cloud of the most frequent artifacts mentioned by the annotators is shown here.
}
\label{fig:annotation_text_wordcloud}
\end{figure}
\begin{figure*}[h!]
\centering
\includegraphics[width=\linewidth]{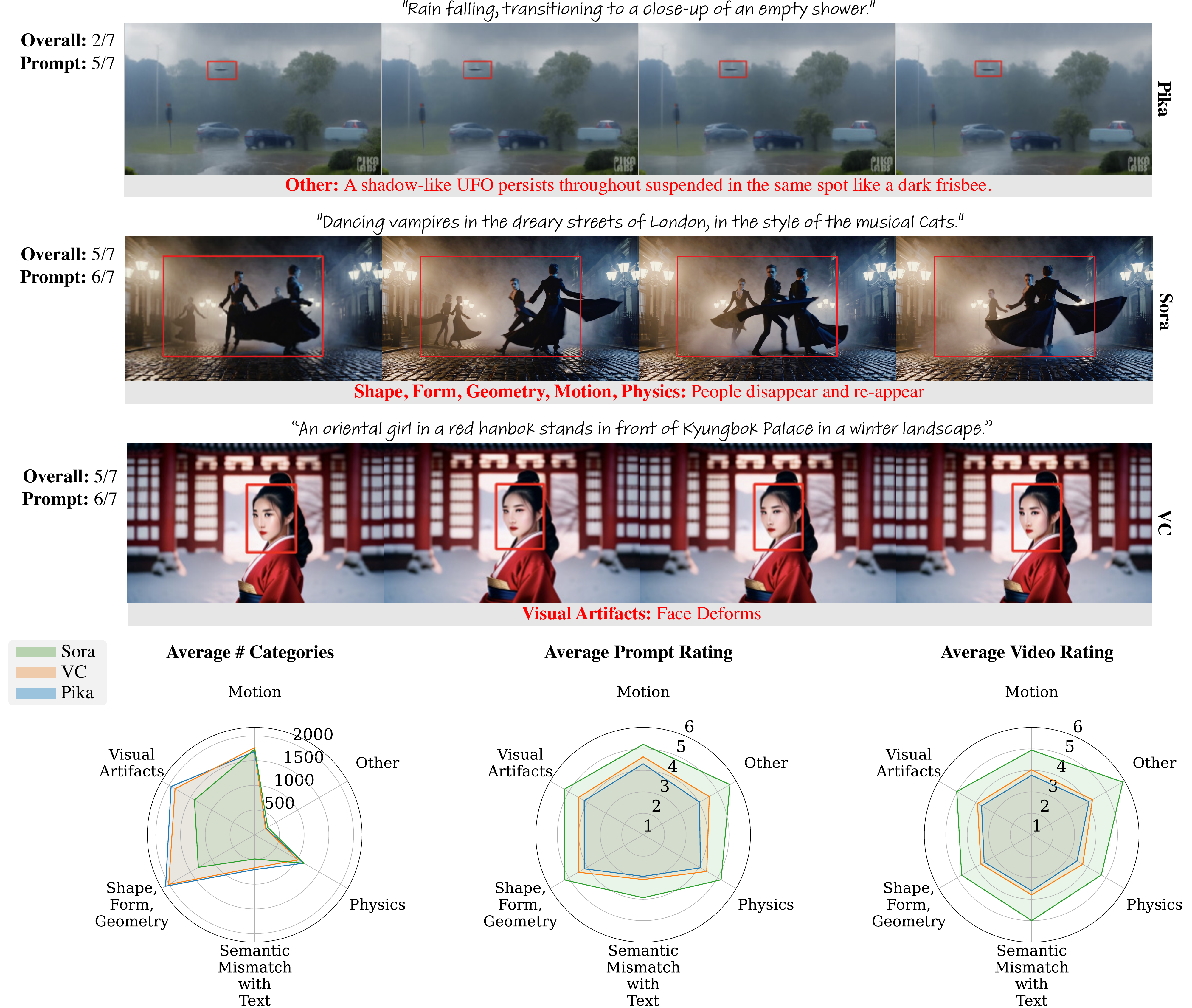}
\caption{
We show example annotated bounding boxes for each model (labeled on the right). 
The bounding boxes are annotated in red, with their artifact category and user-annotated description below the frames.
Video quality (``Overall'') and video-prompt alignment (``Prompt'') are shown to the left.
Summary statistics are shown in the radar plots. 
Specifically, we show statistics for each artifact category, grouped by category count, average video-prompt alignment, and average video quality rating given the user-selected artifact categories. 
This is done for the three models in our dataset: Sora \cite{soramodel}, VideoCrafter2 \cite{chen2024videocrafter2}, and Pika \cite{pikamodel}. See additional examples in the Appendix. 
}
\label{fig:teaser}
\end{figure*}

% %
% %
% \begin{center}
%     \centering
%     \captionsetup{type=figure}
%     \includegraphics[width=\linewidth]{assets/geneva-teaser.pdf}
%     \caption{
% We show example annotated bounding boxes for each model (labeled to the right). 
% The bounding boxes are annotated in red, with their artifact category and user-annotated description below the frames.
% The video quality ("Overall") and video-prompt alignment ("Prompt") are shown to the left.
% Summary statistics are shown at bottom. 
% Specifically, we show statistics for each artifact category, group by category count, average video-prompt alignment and average video quality rating given the user-selected artifact categories. 
% This is done for the three models in our dataset; Sora, VC2, and Pika1.
% \warning{I need someone to get the prompts for these videos. I labeled the video index on the top right of each row.}
% }
% \label{fig:teaser}
% \end{center}%

% \note{
% What prior works have done, and why they are not adequate.
% (Note: this is just high level big ideas. Details should go to a previous work section.)
% }%note
A wide body of research has studied the development of computational metrics to predict visual quality with the goal that the metric correlates well with real subjective responses.
Popular examples of these automatic metrics include SSIM \cite{zhang2018unreasonable}, LPIPS \cite{wang2004image}. 
A number of metrics have also been used to optimize and evaluate text-image generative models. 
For example, the Fréchet Inception Distance (FID) score \cite{heusel2017gans} has been widely adopted to measure generation realism.
However, artifacts that arise from text-video generation may extend beyond what can be captured by these metrics.
An extension of FID to video \cite{unterthiner2019fvd,unterthiner2018towards} has been found to result in poor correlations with subjective judgements in some cases \cite{ge2024content}.
% For instance, incorrect physics may be easily identified by human observers, but not by existing quality metrics. 
To the best of our knowledge, a dataset that comprehensively evaluates and annotates these generative text-video artifacts does not exist, perhaps due to the difficulty in collecting large-scale data of the sort.

% \note{
% What our method has to offer, sales pitch for concrete benefits, not technical details.
% Imagine we are doing a TV advertisement here.
% }%note
In this paper, we present GeneVA, a rich human-annotated dataset for \textbf{gene}rative text to \textbf{v}ideo \textbf{a}rtifacts.
GeneVA offers a representative set of 5,452 text prompts sampled from real-world human text inputs \cite{wang2024vidprom}.
Then, 3 text-video generative models (2 open-source and 1 commercial) were used to generate, with the text prompts as input, 16,356 videos amounting to a sum $52,326$ seconds of content.
In total, 16,451 crowd-sourced annotations of spatio-temporal artifacts was collected.
Annotated artifacts include both categorical selections and free-response textual descriptions (as shown in \Cref{fig:teaser}). Our goal with GeneVA is to provide a robust, challenging dataset to help the research community make systematic progress on understanding complex visual artifacts. The long-term goal is to leverage this understanding to build more reliable and human-centric generative models.
%The collection of this data, in particular, enables analysis of more fine-grained artifact descriptors and allows us to predict human-readable artifact descriptions given AI-generated videos as input.

%As shown in \Cref{fig:teaser}, 
%GeneVA's rich annotations
%, including bounding box locations, categorical ratings and selections, and free-form textual descriptions, 
%enable a number of video artifact analyses for evaluation of AI-generated video. 
As a first step toward this goal, we also leveraged these annotations to develop an interpretable artifact detector pipeline (\Cref{sec:artifact_detector}). In addition to identifying artifact locations, our system also generates human-readable explanations for each artifact, enabling a fine-grained analysis of the generative model's behavior. The artifact detector model achieves 13\% Average Precision ($AP_{0.25}$) and demonstrates strong zero-shot generalization when applied to videos from external models not seen during training. This highlights the robust and generalizable nature of our dataset. 
%We further showcase how GeneVA can fuel a detector model for generative video artifacts\qi{@Patsorn: need to report some numbers to show that it works to some extend.}. 
%\bia{Potential rephrase: The labeled spatio-temporal artifacts include both categorical and free-response textual descriptions, enabling flexible applications, such as a detector model for generative video artifacts (which we explore in Section \ref{sec:artifact_detector})} <3
%\qi{@Patsorn: need to report some numbers to show that it works to some extend.}. 
% \note{
% Our main idea, giving people a take home message and (if possible) see how clever we are.
% }%notet
% \note{
% Our algorithms and methods to show technical contributions and that our solutions are not trivial.
% }%notew

We will release the GeneVA dataset and the trained detector to the public upon acceptance. We envision the dataset could serve as a new benchmark for future text-video generative models, and can launch new research in the direction of artifact reduction and detection in AI-generated videos.
% \note{
% Results, applications, and extra benefits.
% }%note

\section{Related Work}
\begin{table*}[tb]
\caption{
Here, we shows existing datasets that are the most relevant to this work.
We list the total number of models each work used to generate their video dataset, the total number of videos in the dataset, the total number of human ratings, whether participants had to describe artifacts, and whether participants localized artifacts by annotating bounding boxes. 
Notably, our dataset is the only one to include bounding boxes that specify the artifact locations and text annotations that describe their nature.
% This provides greater insight into the visual features that tend to create artifacts in generated videos.
}
\resizebox{\linewidth}{!}{
\begin{tabular}{ l | c | c | c | c | c  }
\hline
Dataset & Total Model Count & Total Video Count & Total Human Rating Count & Description? & Bounding Box? \\ 
\hline
T2V \cite{chivileva2023measuring} & 5 & 1,005 & 48,240 & \xmark & \xmark \\
Vbench \cite{huang2024vbench} & 4 & $\sim$1,700 & -- & \xmark & \xmark \\
FETV \cite{liu2023fetv} & 4 & 2,613 & 28,116 & \xmark & \xmark \\
EvalCrafter \cite{liu2023evalcrafter} & 5 & 2,500 & 8,647 & \xmark & \xmark \\
% VideoReward \cite{liu2025improving} & & & & & \xmark \\
VideoScore \cite{he2024videoscore} & 11 & 37,600 & -- & \xmark & \xmark \\
\hline
\textbf{GeneVA} & 3 & 16,356 & 16,356 & \cmark & \cmark \\
\hline
\end{tabular}
}

\label{table:prior-art}
\end{table*}

In this section, we summarize prior datasets on text-video generation.
We make an effort to focus on datasets that collect human quality assessments of these videos.
A summary of the most relevant datasets is tabulated in \Cref{table:prior-art}.

\subsection{Datasets for Annotating Generative Videos}
The proliferation of generative AI has created a need for the quality assessment of AI-generated images and videos \cite{liu2024ntire}.
Numerous datasets have been made available that include human assessments of AI-generated images \cite{wu2023human,xu2023imagereward, kirstain2023pick,chen2023exploring}.
Fewer datasets of this kind have been proposed for text-video content. 
\citet{chen2024gaia} collected quality data for 9,180 videos, focusing specifically on video-action pairs.
VideoReward \cite{liu2025improving} and IPO \cite{yang2025ipo} collected pairwise comparison data for content generated by a number of different video generation models.
\citet{chivileva2023measuring} and \citet{huang2024vbench} collected human quality ratings and measured their alignment with quality metrics.
The VideoScore \cite{he2024videoscore}, FETV \cite{liu2023fetv}, and EvalCrafter \cite{liu2023evalcrafter} datasets collected human rating across various aspects (temporal consistency, text alignment, etc.) on a large set of generated videos.
None of these works, however, include bounding boxes on artifacts or free-form text descriptions of them.

\subsection{Artifact Detection}
The detection and localization of artifacts in text-image and text-video content can facilitate identification of AI-generated content.
This could be useful for detecting fraud or mitigating the spread of misinformation, for example \cite{prashnani2024generalizable,rana2022deepfake}.
Several datasets are available for annotated bounding boxes in text-image content \cite{cao2024synartifact,wang2024detecting,wang2025generated,fang2024humanrefiner}.
In this work, we focus on the localization of artifacts in AI-generated videos.
Very few works have collected a dataset of this sort for text-video generation; one such example is for the task of deepfake detection by \citet{hondru2025exddv}.
No prior works that we are aware of provide annotated bounding boxes for general text-video content.

\section{The GeneVA Dataset}

In this section, we describe how the text-video dataset is collected and how human annotators label each video.

\subsection{Synthetic Video Collection from Text-to-Video Models}
\begin{figure*}[t]
\centering
\subfloat[original embedding space]{\includegraphics[height=5.2cm,trim={0 1.2cm 0 1.5cm},clip]{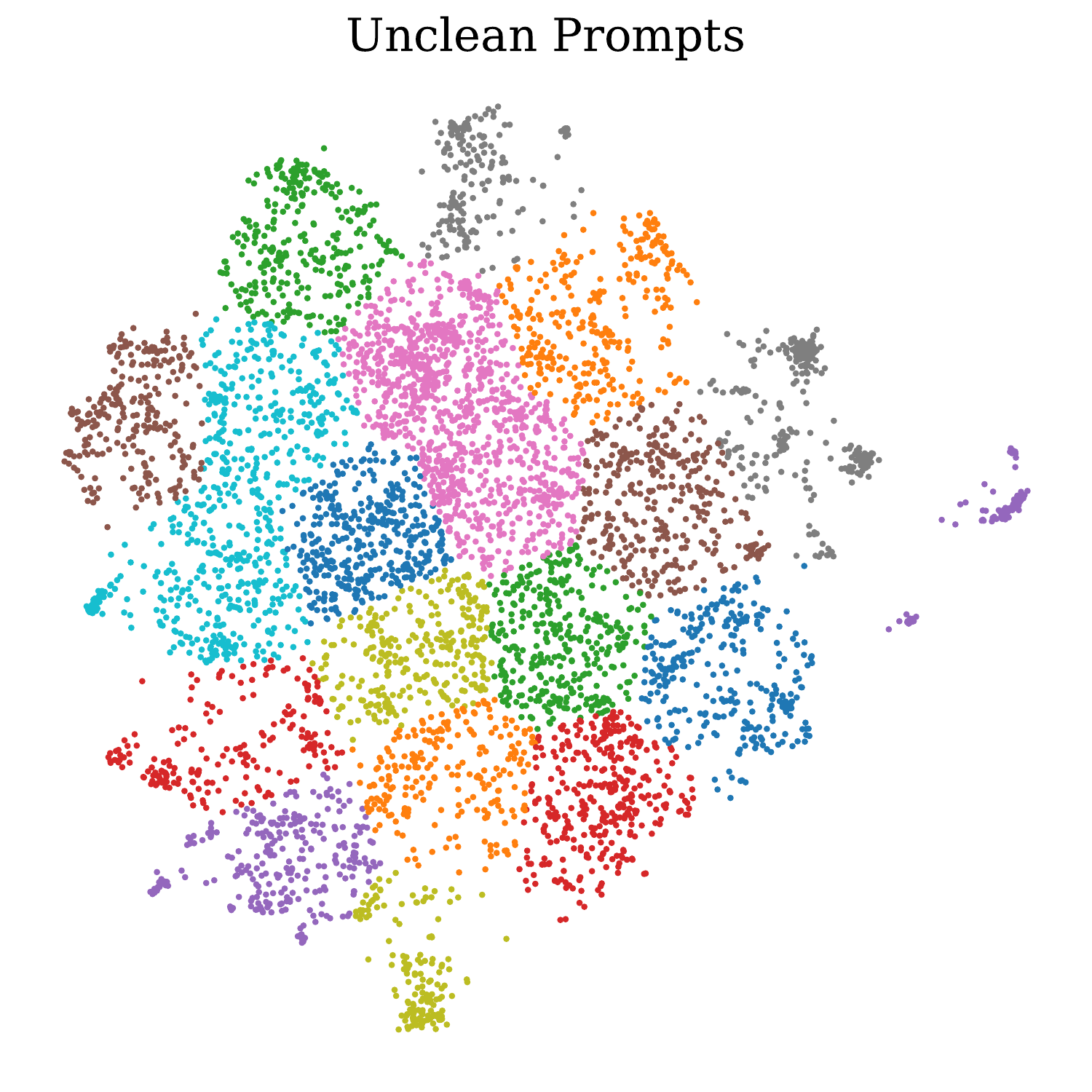}\label{fig:prompt-sampling:original}}
\subfloat[cleaned embedding space]{\includegraphics[height=5.2cm,trim={0 1.2cm 0 1.5cm},clip]{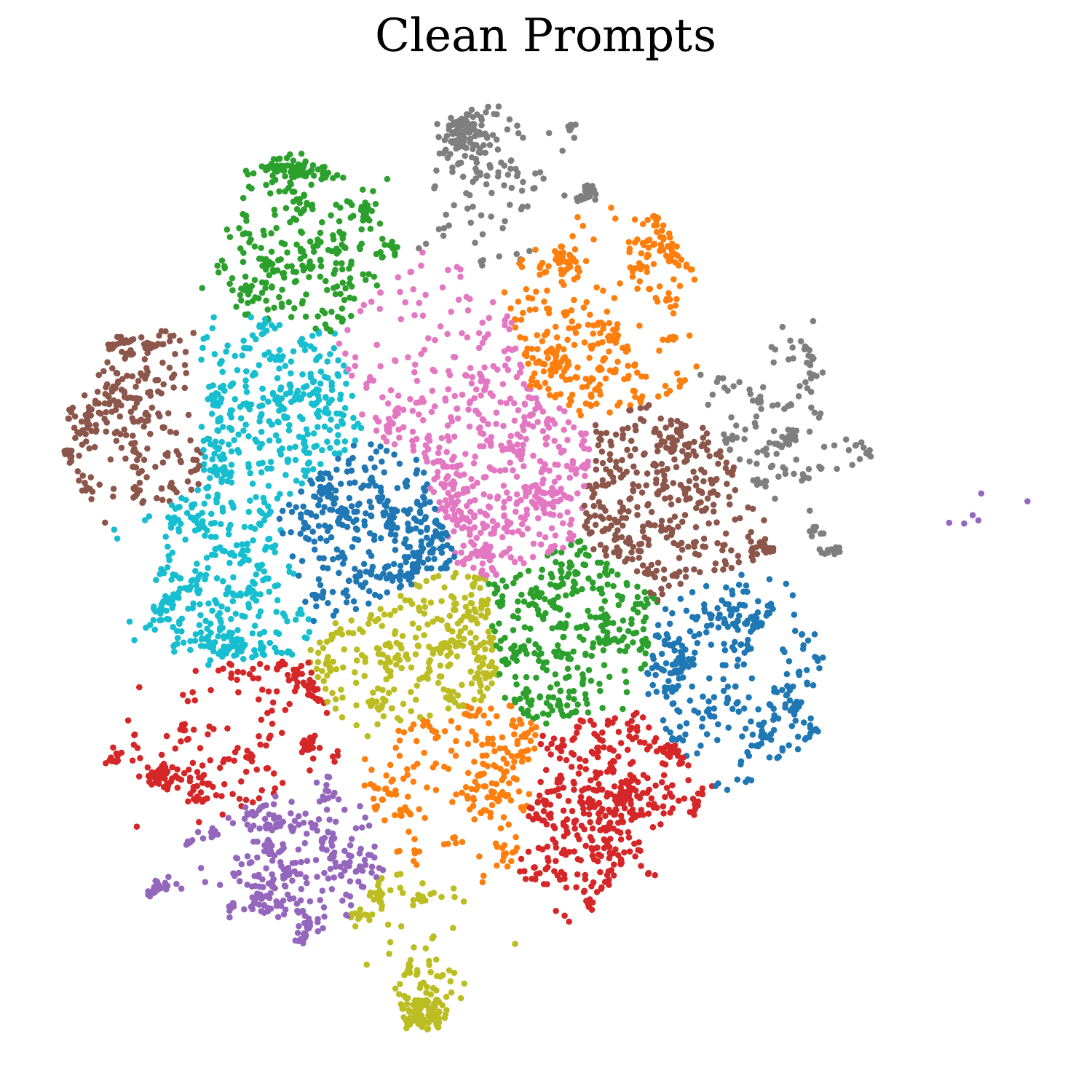}\label{fig:prompt-sampling:sampled}}
\subfloat[language prompts]{\includegraphics[height=5.2cm]{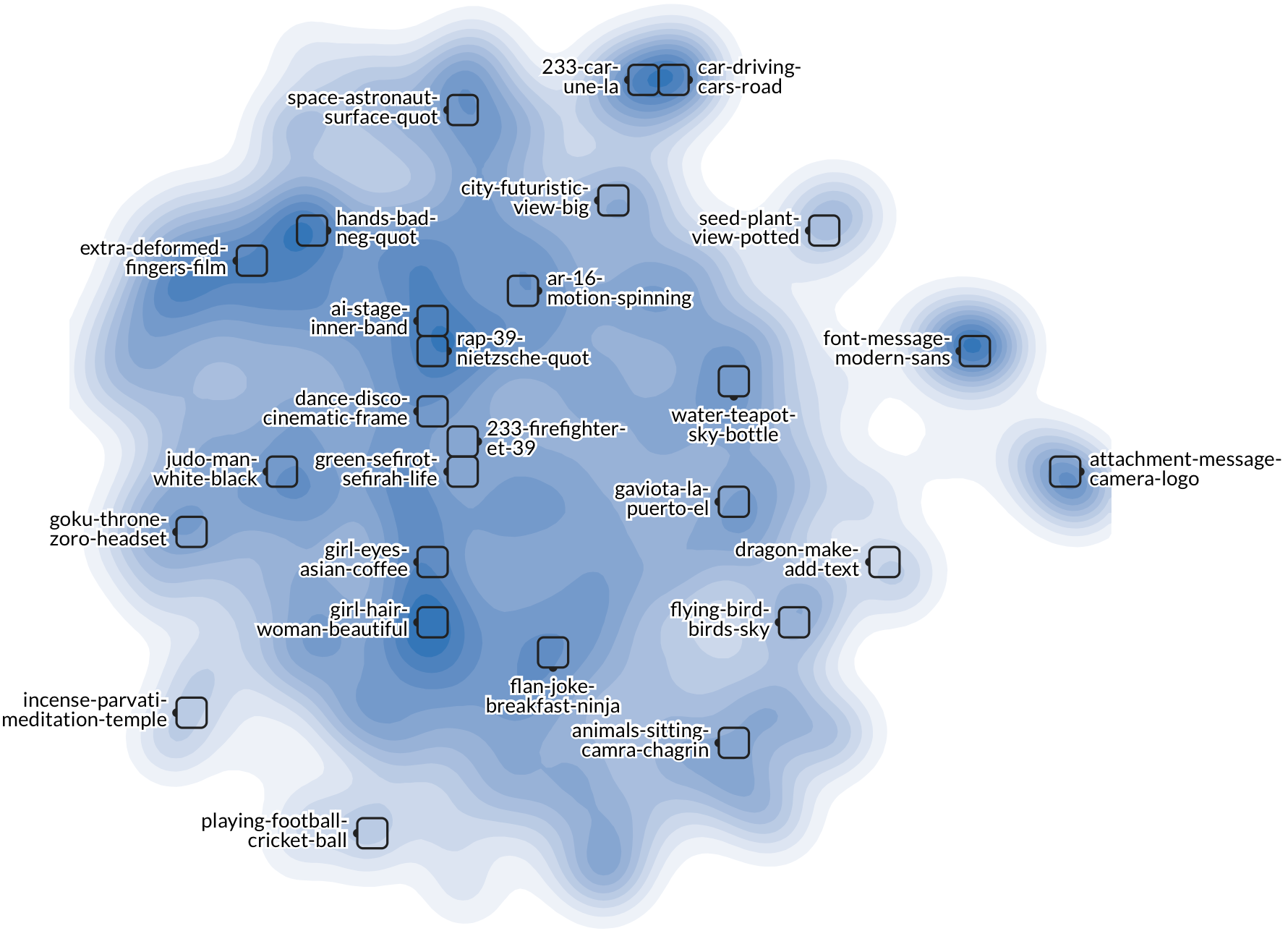}
\label{fig:wizmap}}
\caption{
\subref{fig:prompt-sampling:original}/\subref{fig:prompt-sampling:sampled} visualize the embedded feature space before and after our cleaning procedure.
We note that the two plots look very similar.
% It can be seen that our cleaned prompts uniformly covers the feature space to comprehensively cover natural semantic groups that were present in the original, uncleaned prompts used to generate videos.
Plots with different cluster counts are displayed in the Appendix.
% \kenny{Are these plots for before/after we cleaned w/ gemini?}
\protect\subref{fig:wizmap} We visualize the embedding space of prompts used in our dataset via Wizmap \cite{wang2023wizmapscalableinteractivevisualization}. Darker colors represent denser regions, which are labeled with their descriptors. 
% It can be seen that the embedding distance represents language, thus semantic similarities. \qi{feel free to make this fig single column if we are out of space.}
}
\label{fig:prompt-sampling}
\end{figure*}

To effectively analyze artifacts in text-to-video generation, we first require a robust and representative dataset of synthetic videos. Our goal is to capture realistic scenarios and diverse content typically encountered in real-world text-to-video generation use cases. 

We constructed our dataset by leveraging the large-scale VidProM dataset \cite{wang2024vidprom}, which contains 1.67 million unique prompts derived from real-world user interactions (scraped from Pika Discord servers) and corresponding videos from multiple text-video generation models. To extract a representative subset that maintains the distributional properties of this extensive collection, we employed kernel herding \cite{chen2012kherding} on 3,072-dimensional embeddings of the prompts, which were generated using OpenAI's text-embedding-3-large model \cite{openaiembedding}. We sampled 6,000 prompts using this approach, with the goal of ensuring fair coverage across the semantic space of prompts, avoiding sampling bias while preserving the diversity of the original dataset. \Cref{fig:wizmap} visualizes the distribution of our selected prompts. More details are available in the Appendix.
% Kenny: I commented out the link to the wizmap cuz it includes identifying info in the url. I doubt anyone would see that, but just in case.
% \footnote{an interactive version is available \href{https://poloclub.github.io/wizmap/?dataURL=https%3A%2F%2Fmariabeatrizsilva.github.io%2Fprompt-embeddings%2Fdata.ndjson&gridURL=https%3A%2F%2Fmariabeatrizsilva.github.io%2Fprompt-embeddings%2Fgrid.json}{online}}.
% \Cref{fig:prompt-sampling:original,fig:prompt-sampling:sampled} visualizes the embedded prompt distributions before and after the sampling, respectively.

To maximize diversity in terms of generated videos and potential artifacts, our dataset incorporates videos from three recent generative text-video models: Sora \cite{soramodel}, Pika \cite{pikamodel}, and Videocrafter2 (VC) \cite{chen2024videocrafter2}. 
Videos generated by Pika and VC are directly extracted from the VidProM dataset \cite{wang2024vidprom}. We supplemented these with newly generated Sora videos to capture the current state-of-the-art performance and ensure our analysis reflects the most recent advances in text-to-video generation. For the Sora model, we generated 5-second videos for each sampled prompt. Some prompts from our original prompt sampling were excluded due to moderation restrictions of the Sora model as well as our own manual review to ensure appropriate video content. This careful curation resulted in a final dataset of 5,452 unique prompts and 16,356 synthetic videos across the three models, yielding a total of $52,326$ seconds of diverse video content.

% \niall{i tihnk we are going to need strong justification for why we dont use models like google's new flow model. i expect reviewers will ask about this, so we should try to preempt this question in this section. Patsorn suggests: we could angle it as supporting the opensource model that don't have as many data as large company current opensource model move really slowly and all large company never release their model}
% \bia{One thing we might want to figure out how to mention was how we chose our size/the challenge sampling the dataset posed. A major part of getting the videos (at least for me) was the fact we wanted a smaller, yet representative-sample of a *huge* dataset. I tried to address it a little below, but I feel like providing context (on how/why we chose the smaller size) might be useful? And this feels linked to the need to generate our own videos/justify the use of VidProM}

\subsection{Human Annotations}
\label{sec:annotation}
\begin{figure}[t]
\centering
\includegraphics[width=\linewidth]{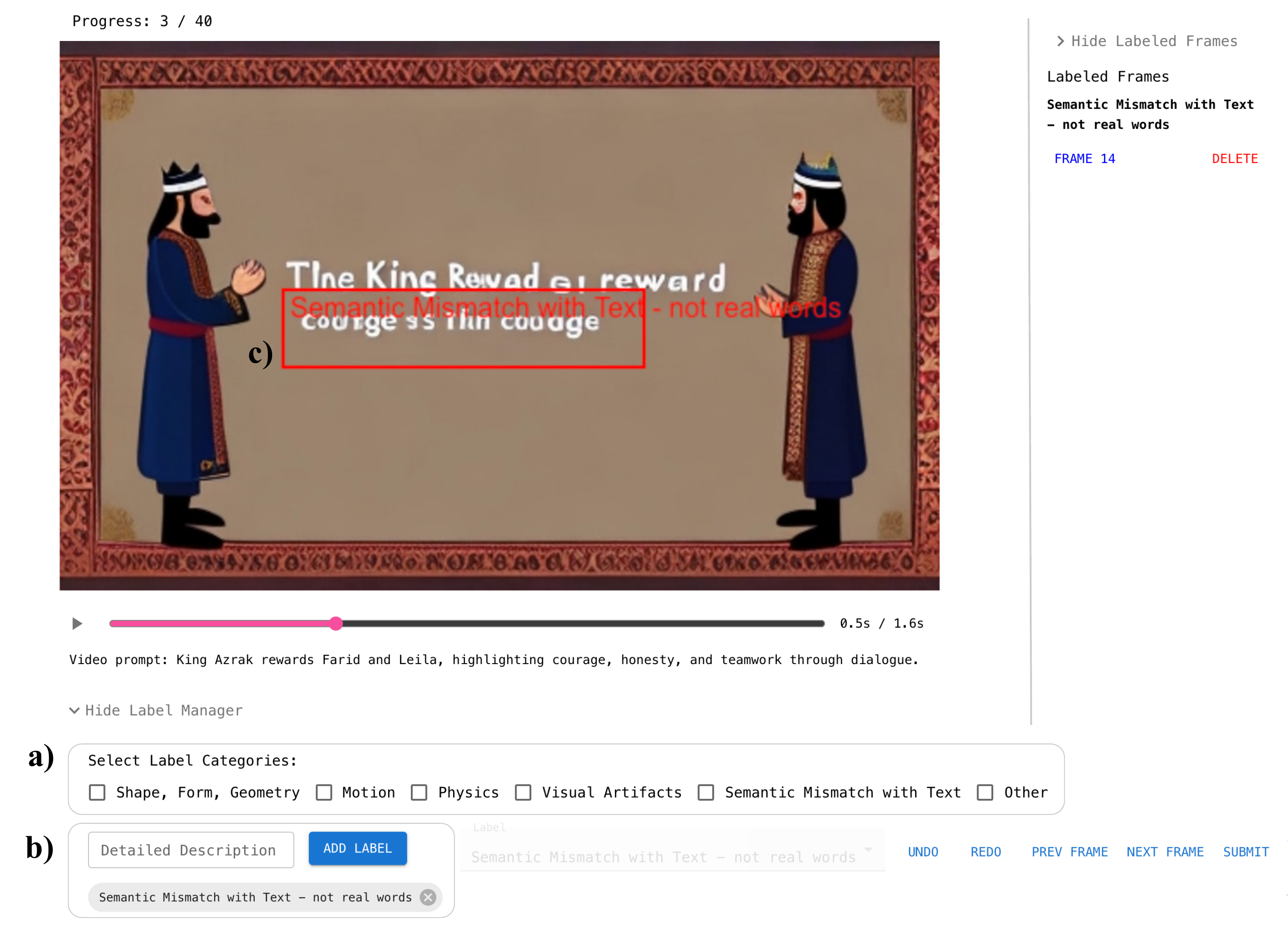}
\caption{
A screenshot of our video annotation interface.
Users annotate artifacts by (a) selecting a label category, (b) describing the artifact in additional detail (free response), and (c) drawing a bounding box around the artifact.
Bounding boxes for the same labeled artifact are interpolated across multiple frames as long as the user draws an initial and final keyframe bounding box for the artifact.
% Users can add up to 5 labels.
Annotated bounding boxes are listed in the right side column (enumerated per-frame) to allow for easy understanding of which frames have been annotated and simple editing if users want to change their boxes before submitting.
}
\label{fig:annotation}
\end{figure}

%After collecting the text prompt-video pairs, our goal is to gather real user annotations of them, including of their quality ratings, artifact labels, and artifact bounding box localizations.
Our goal is to gather human annotations for our curated text-and-video pairs. Specifically, we aim to collect participants' quality ratings of each video, as well as annotations of artifacts, including their category, description, and location. To gather this data, we run a large-scale crowdsourced study described below.

\paragraph{Participants}
We recruited study participants through the online crowdsourcing platform \textit{Prolific}. All study protocols were approved by an institutional review board (IRB), and subjects were compensated at a rate of \$10/h.

\paragraph{Study Protocol}
A screenshot of our web-based annotation tool is shown in \autoref{fig:annotation}.
The study was conducted via a web browser, and run on the participant's computer screen. 
We did not enforce any requirements on users' display specifications to better reflect real-world video viewing conditions. 
%-- we believed that allowing users to view text-video content on diverse viewing conditions is more representative of a real-world scenario.
Subjects were shown a 3-minute tutorial video on example artifact labeling, controls, and procedures before proceeding with the annotation task. 
Each video was played in its entirety before participants were allowed to begin the annotation phase. 
During the annotation phase, participants are shown the video and its corresponding text prompt. For clarity, we provide the cleaned versions of the original VidProM prompts that preserves the original's semantic intent (see \Cref{fig:prompt-sampling:original,fig:prompt-sampling:sampled} and Appendix for more details). 
%(see Appendix for details).
%Qualitatively, we find that original and cleaned prompts align well in intent (see \Cref{fig:prompt-sampling:original,fig:prompt-sampling:sampled}).
Participants annotated artifacts by selecting a label category from a pre-defined set (\Cref{fig:annotation} (a)) and describing the artifact in additional detail (\Cref{fig:annotation} (b)). 
Artifact categories are selected to cover the most representative artifact types in AI-generated video, and includes 1) Shape, Form, Geometry, 2) Motion, 3) Physics, 4) Visual Artifacts, and 5) Other. 
Participants were free to scrub through the video timeline to jump to specific frames of the video.
For each label, participants had the choice to draw a bounding box around the artifact (\Cref{fig:annotation} (c)). 
Bounding boxes for the same labeled artifact are interpolated across multiple frames. 

To encourage users to prioritize reporting salient artifacts rather than over-scrutinizing the videos, we set a limit of five artifact annotations per video. This ensures that our data primarily captures artifacts most likely to be noticed by a typical user.
%There was a limit of five labels per video to ensure participants did not over-scrutinize the videos.
%This could lead users to report artifacts that, although present, are not noticeable to most users under typical viewing conditions.
%Furthermore, by limiting the number of bounding boxes to five, users are naturally encouraged to report the most salient artifacts contributing to degraded video quality.
Once the user submits their labeled artifacts, they rate the video's visual quality and its alignment with the text prompt. 
Users submitted ratings on two separate 7-point Likert scales (where 1 being the lowest quality and 7 the highest).

\section{Dataset Characteristics}
% \warning{need to include discussion of Fig. 3).}

\begin{figure}[t]
\centering

\subfloat{\includegraphics[width=.49\linewidth]{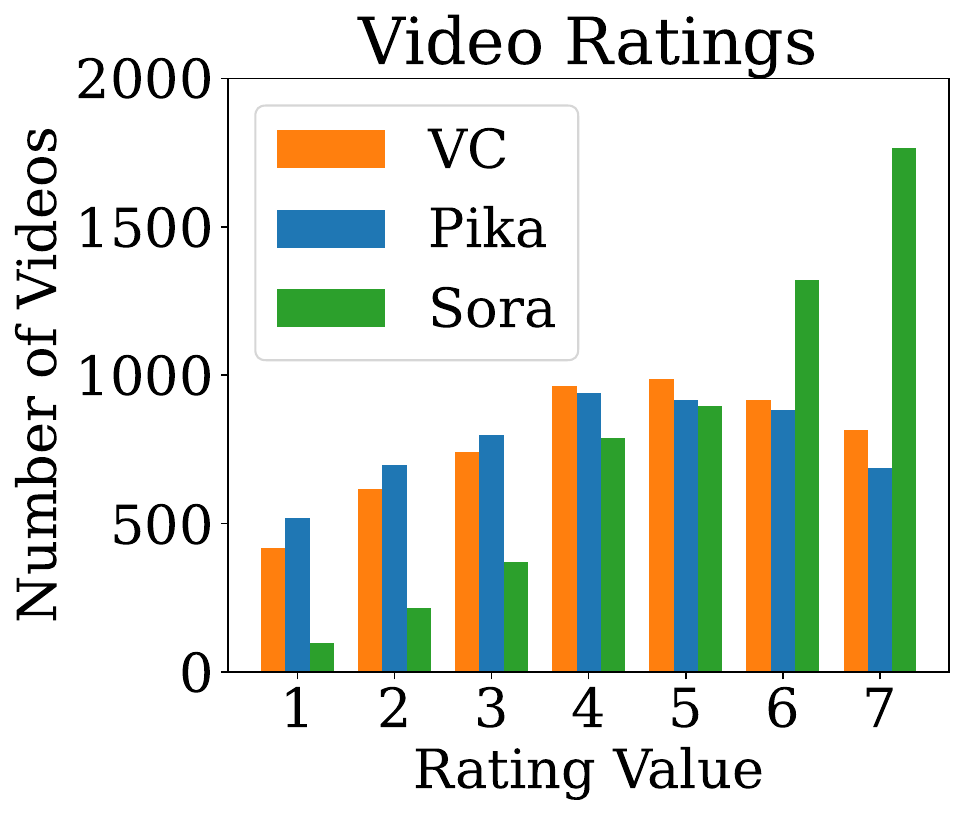}\label{fig:video_rating_distribution}}
\subfloat{\includegraphics[width=.49\linewidth]{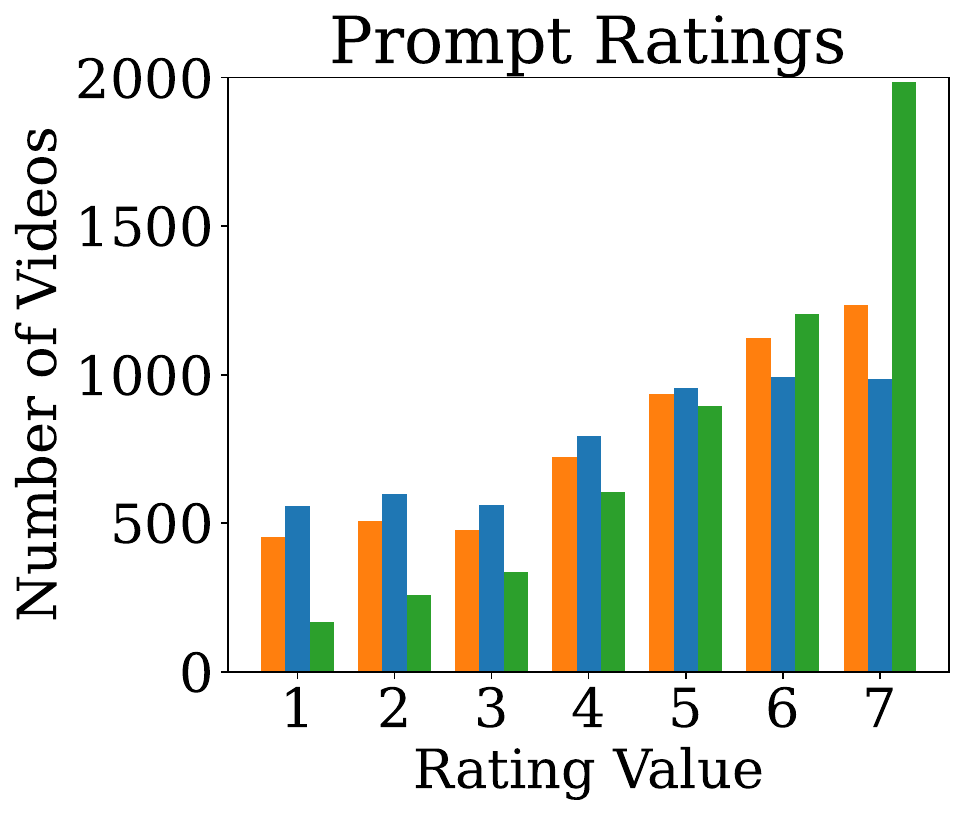}\label{fig:prompt_rating_distribution}}
\caption{
Distributions of user-submitted scores for each video's (left) visual quality and (right) alignment between text prompt and generated content.
Both the video and text quality scores are notably higher for Sora videos compared to the videos generated by VC and Pika, which share very similar score distribution shapes.
}
\label{fig:ratings}
\end{figure}
\begin{figure}[t]
\centering
\includegraphics[width=\linewidth]{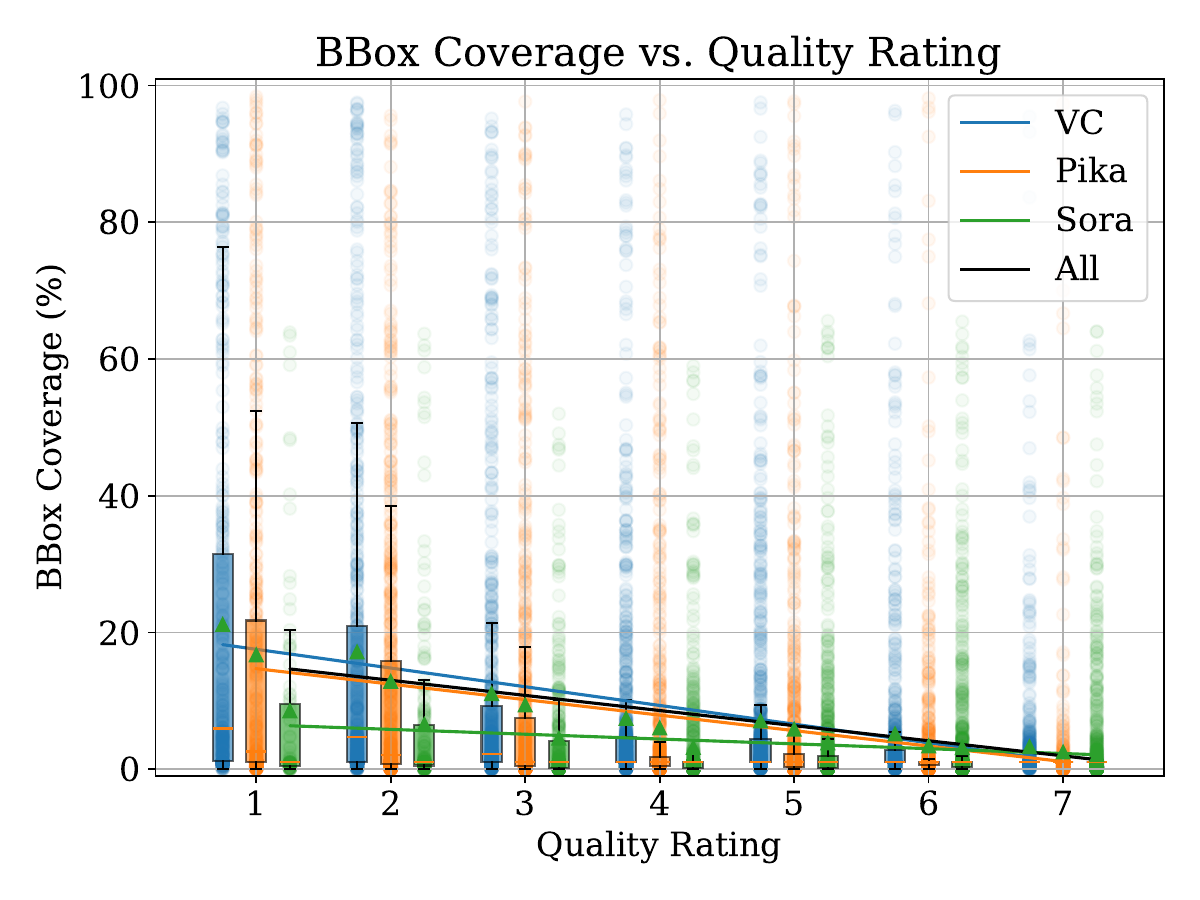}
\caption{
Box plots showing the distribution of spatio-temporal bounding box coverage across all videos, grouped by generative model and video quality score. 
Whiskers represent 95\% confidence intervals.
We also show the regression fits that denote the correlation between video quality and bounding box coverage. 
% All correlations were statistically significant, indicating that the prevalence of bounding boxes in a video predicts users' subjective quality ratings. Since Sora videos consistently have fewer artifacts in general, the correlation between bounding box coverage and quality rating is weaker for Sora compared to VC and Pika videos. 
The black line shows the regression for the pooled data of all models.
% \warning{
% Data summary:
% VC2 → r=-0.27, p=0.000, slope=-2.74, intercept=20.96 | 
% PIKA1 → r=-0.25, p=0.000, slope=-2.27, intercept=16.98 | 
% SORA → r=-0.13, p=0.000, slope=-0.71, intercept=7.06 | 
% ALL → r=-0.26, p=0.000, slope=-2.21, intercept=16.88
% }
}
\label{fig:bbox_coverage_vs_quality}
\end{figure}

% \Cref{fig:ratings} shows the distribution of ratings throughout all participants; \Cref{fig:annotation_text_wordcloud} 
% %visualizes the text-labeled artifacts in language embedding space.
% shows a word cloud visualization of the most common words from the text-labeled artifacts.
%
To better understand what kinds of video content our subjects annotated, in this section, we analyze and summarize various properties of the AI-generated videos.
A base summary of the free-form artifact descriptions users submitted is visualized in \Cref{fig:annotation_text_wordcloud}, which shows the most common words used to describe artifacts.
The distribution of video quality and text prompt-alignment ratings is shown in \Cref{fig:ratings}, separated by model.
Interestingly, text prompt alignment scores were right-skewed, but seem normally-distributed for video quality (except for Sora).

\begin{figure*}[t]
\centering
\subfloat[average video rating]{\includegraphics[height=5.4cm,trim={0 0 0 1.25cm},clip]{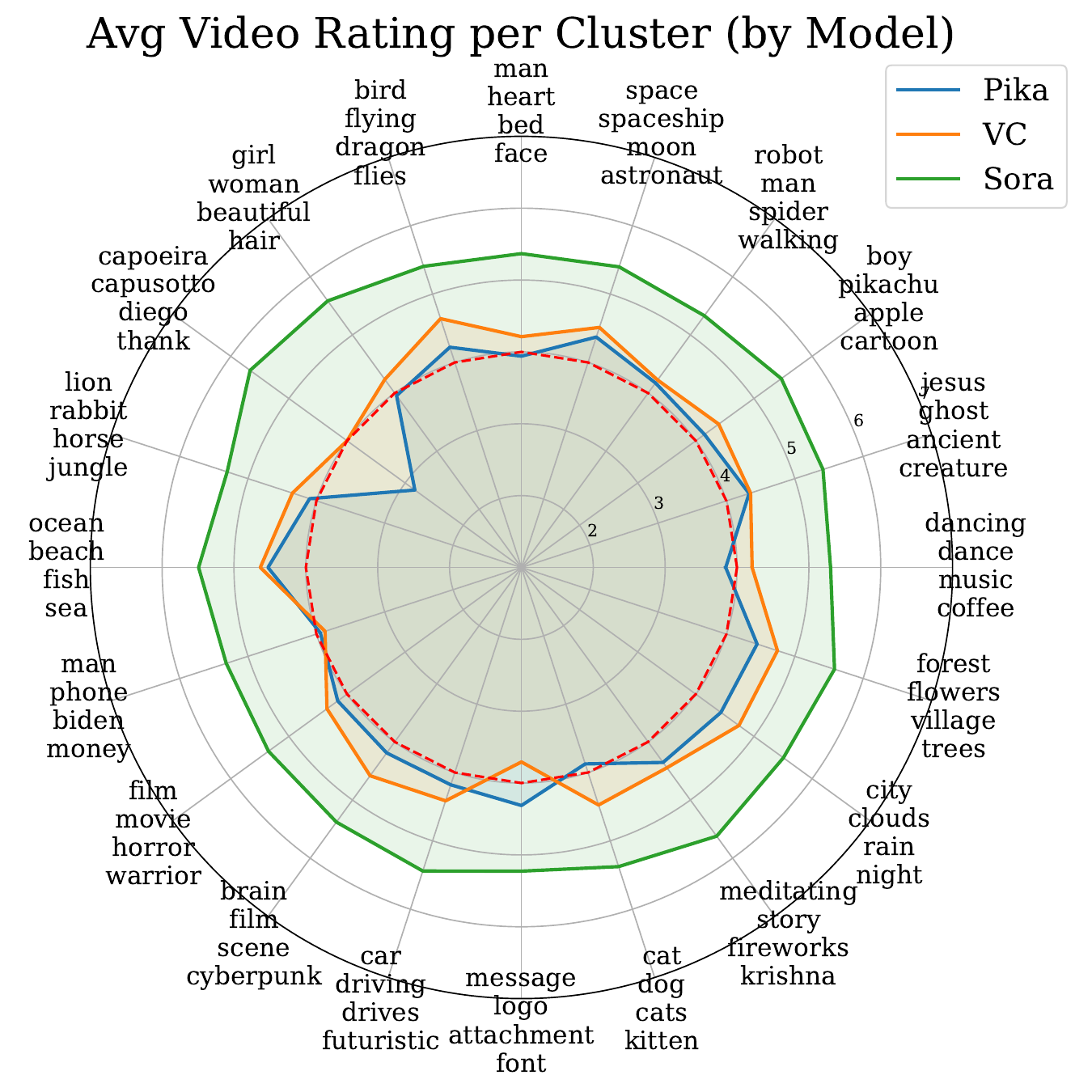}\label{fig:all-stats-vid-rating}}
\hspace{0.1em}
\subfloat[average prompt rating]{\includegraphics[height=5.4cm,trim={0 0 0 1.25cm},clip]{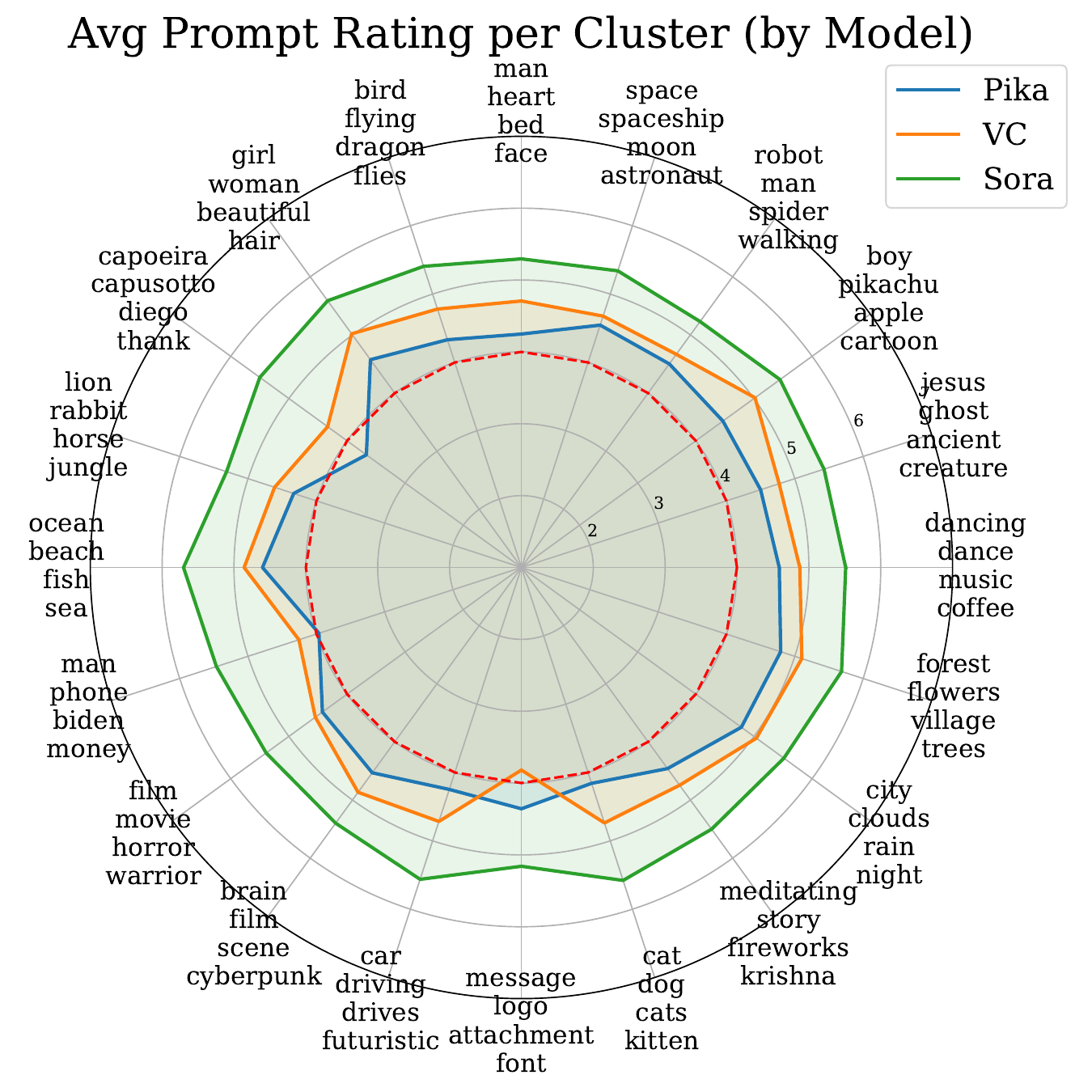}\label{fig:all-stats-prompt-rating}}
\hspace{0.1em}
\subfloat[average \# artifacts]{\includegraphics[height=5.4cm,trim={0 0 0 1.25cm},clip]{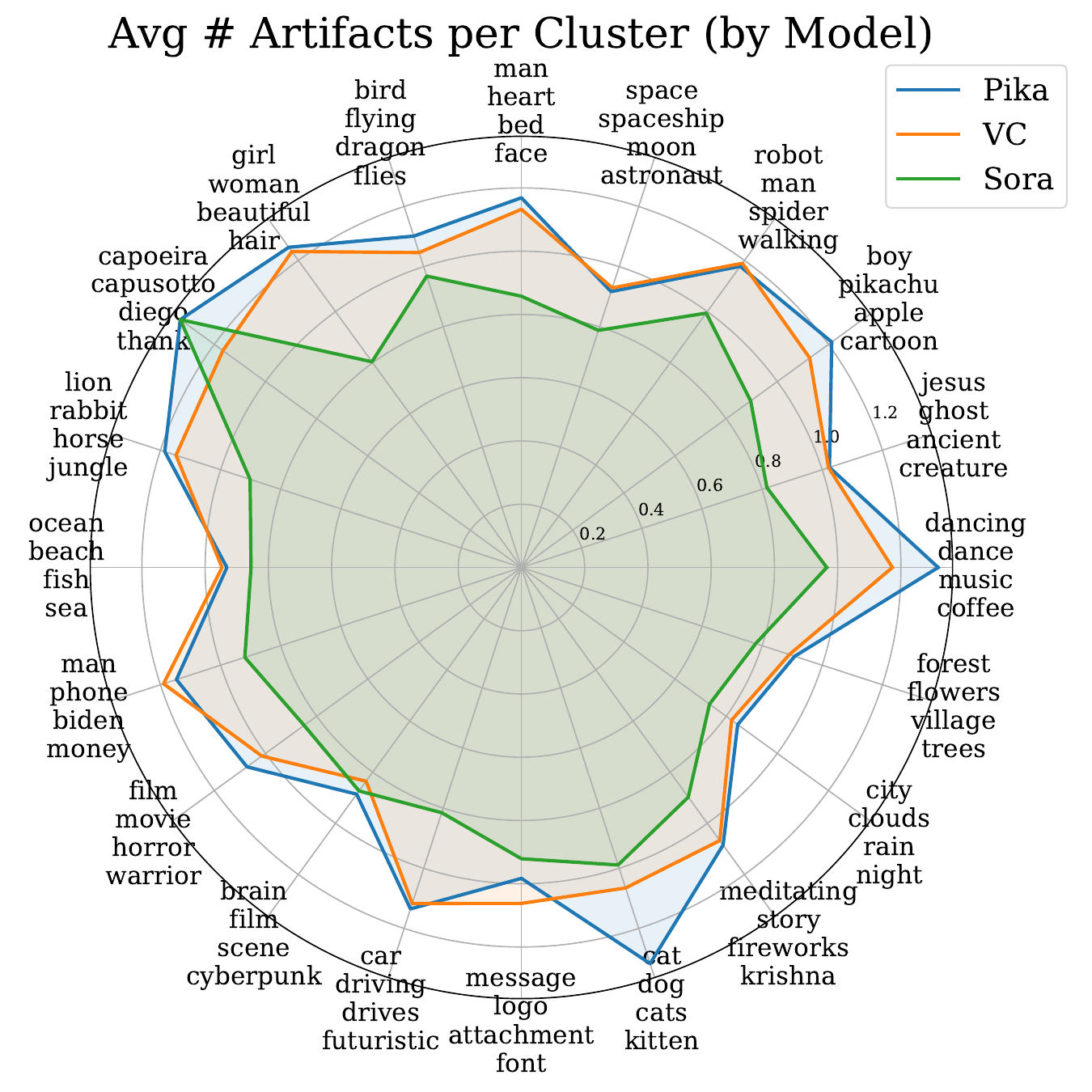}\label{fig:all-stats-artifact-count}}

\vspace{1ex}

\subfloat[average \# bboxes]{\includegraphics[height=5.4cm,trim={0 0 0 1.25cm},clip]{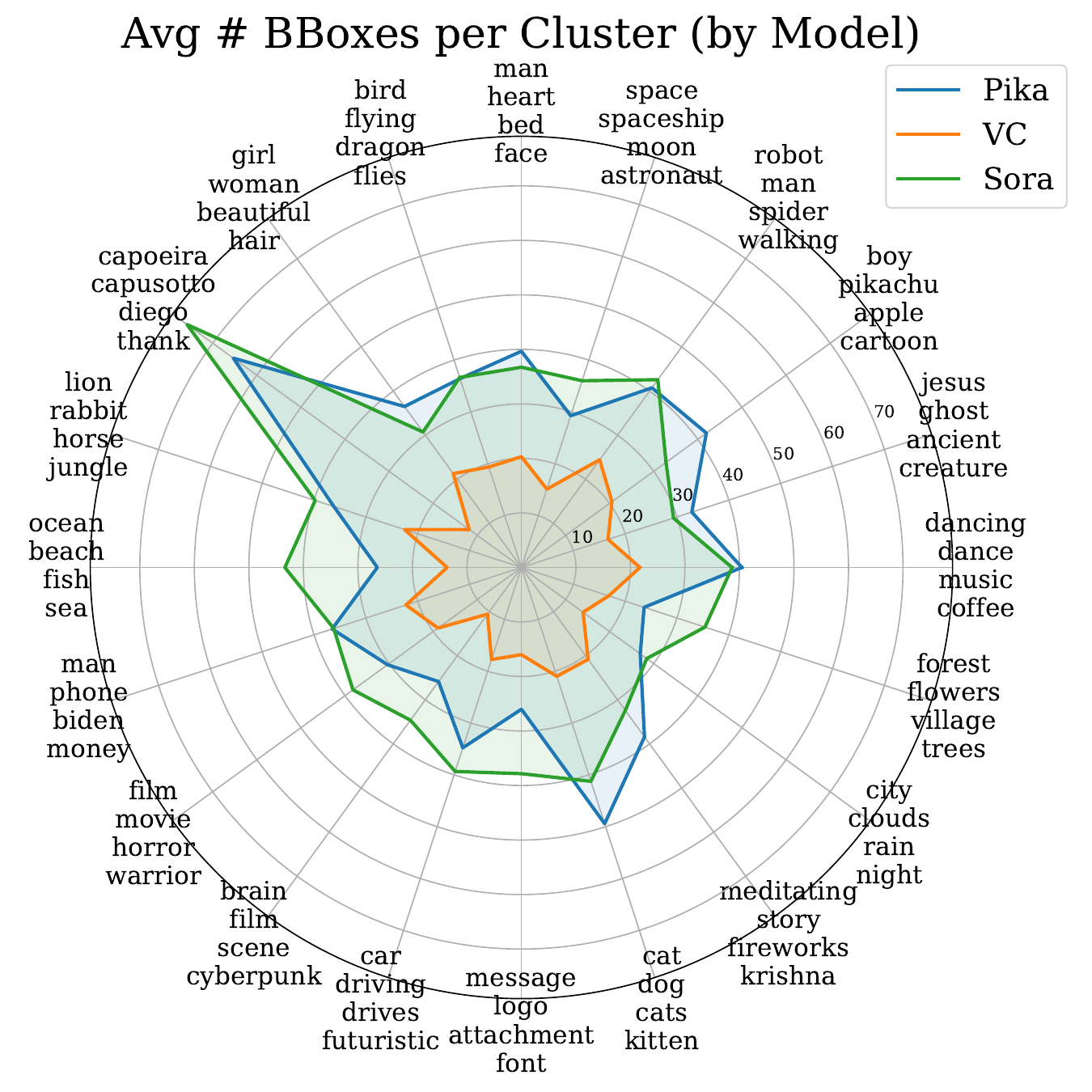}\label{fig:all-stats-bbox-count}}
\hspace{0.1em}
\subfloat[average total bbox coverage]{\includegraphics[height=5.4cm,trim={0 0 0 1.25cm},clip]{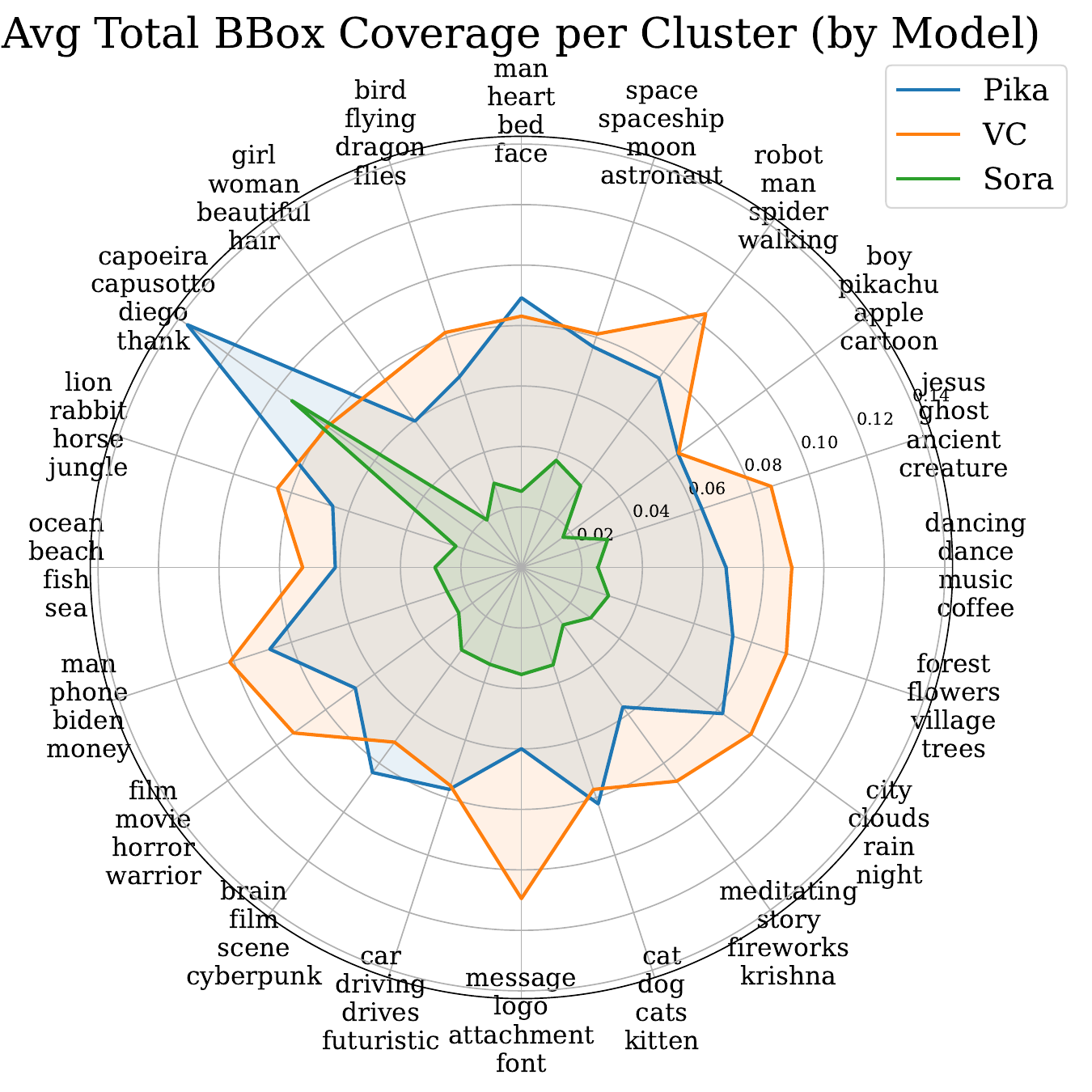}\label{fig:all-stats-bbox-coverage}}

\caption{
Dataset statistics grouped by the top 20 semantic clusters and grouped by model.
The example words for each cluster are the top 4 words in that cluster (zoom for better view). The dashed red line in subplots \subref{fig:all-stats-vid-rating} and \subref{fig:all-stats-prompt-rating} indicate the midpoint of the 7-point rating scale.
}
\label{fig:dataset-statistics}
\end{figure*}

Video summary statistics are shown in \Cref{fig:dataset-statistics}, grouped into 20 different semantic clusters (results for additional cluster counts are displayed in the Appendix).
To interpret the discovered text prompt clusters, we analyzed their content using TF-IDF and embedding-based proximity. 
First, we filtered for prompts from clean sources and computed their TF-IDF representations using a vocabulary of the top 10,000 words after removing English stopwords. 
For each cluster, we calculated the mean TF-IDF scores and compared them to global means across all prompts to identify cluster-specific keywords. 
A log-ratio-based scoring function was applied to highlight words that are both frequent within a cluster and relatively rare globally. 
The top 4 scoring words were selected as representative keywords for each cluster.

Based on the trends shown in \autoref{fig:dataset-statistics}, we observe that Sora generally outperforms Pika and VC, scoring higher on average prompt and video quality ratings, as well as lower on the average number of artifacts and total bounding box coverage.
Furthermore, VC and Pika video ratings hover around the midpoint of the 7-point rating rating scale (red dashed line in \Cref{fig:all-stats-vid-rating,fig:all-stats-prompt-rating}, while Sora is much higher than the midpoint rating.
Interestingly, VC scores notably lower on the average number of bounding boxes per cluster.

\begin{table}[t!]
\centering
\caption{
We show per-model characteristics, comparing videos having a semantic match with the text prompt and those with a mismatch. 
% For each model, we split the videos into two groups: videos with at least artifact labeled in the ``Semantic Mismatch'' category and videos without any artifacts labeled in the ``Semantic Mismatch'' category. 
% Compared to videos with a semantic mismatch, semantic match videos consistently scored higher in average video quality scores (Quality), average text prompt alignment scores (Prompt) and consistently scored lower in average number of artifacts (per video) and average number of bounding boxes (across all frames per video). 
These results highlight the importance of ensuring that generated videos are aligned with the semantics of the text prompts used to generate them.
% , as it leads to consistently higher overall video quality.
}
\resizebox{\linewidth}{!}{
\begin{tabular}{lcccccc}
\hline
\textbf{Model} & \textbf{Category} & \textbf{Quality} & \textbf{Prompt} & \textbf{\# Artifacts} & \textbf{\# BBoxes} & \textbf{\# Samples} \\
\hline
\rowcolor[gray]{0.9} Pika & Semantic Mismatch & 3.604 & 2.938 & 1.59 & 48.12 & 704 \\
\rowcolor[gray]{0.9} Pika & Semantic Match      & 4.268 & 4.677 & 1.02 & 32.25 & 4743 \\
VC   & Semantic Mismatch & 3.799 & 3.084 & 1.54 & 27.29 & 667 \\
VC   & Semantic Match      & 4.453 & 4.971 & 0.99 & 17.75 & 4785 \\
\rowcolor[gray]{0.9} Sora  & Semantic Mismatch & 4.998 & 3.933 & 1.48 & 76.00 & 489 \\
\rowcolor[gray]{0.9} Sora  & Semantic Match      & 5.460 & 5.600 & 0.81 & 32.25 & 4963 \\
\hline
\end{tabular}
}
\label{tab:model_mismatch_metrics}
\end{table}

Since we are studying the quality of text-to-video generative models, we also need to consider the extent to which the semantic alignment between generated videos and their input text prompts influences users' subjective rating of video quality.
\autoref{tab:model_mismatch_metrics} shows the average quality score for videos, grouped by model and whether or not users reported a semantic match between the video content and the text prompt.
Unsurprisingly, the average quality and prompt alignment scores are lower for semantic mismatches.
Furthermore, videos that had a semantic mismatch received more reports of artifacts (`\# Artifacts') and had more frames on average with at least one bounding box annotation (`\# BBoxes').
%
% Data summary:
% VC2 → r=-0.27, p=0.000, slope=-2.74, intercept=20.96 | 
% PIKA1 → r=-0.25, p=0.000, slope=-2.27, intercept=16.98 | 
% SORA → r=-0.13, p=0.000, slope=-0.71, intercept=7.06 | 
% ALL → r=-0.26, p=0.000, slope=-2.21, intercept=16.88

Finally, we inspected the correlation between video quality scores and the average spatio-temporal coverage of bounding boxes across all videos.
Spatio-temporal coverage was defined as the percentage of frames and pixels that were covered by at least one bounding box for a given video.
\autoref{fig:bbox_coverage_vs_quality} shows the box plot distributions of bounding box coverages for each video, as well as the regression lines for the correlation between video quality scores and bounding box coverage.
Unsurprisingly, we see a weak but statistically significant inverse correlation between quality scores and bounding box coverage for all models.
As the quality ratings of videos increase, the amount of bounding box coverage decreases, indicating that higher-rated videos have fewer perceived artifacts.
The strongest correlation (Pearson) between scores and coverage was found in the VC model ($r=-0.27, p<0.0001$)
% , \text{slope}=-2.74, \text{intercept}=20.96$)
, with Pika having the next strongest correlation ($r=-0.25, p<0.0001$),
% \text{slope}=-2.27, \text{intercept}=16.98$),
and Sora with the weakest correlation ($r=-0.13, p<0.0001$).
% \text{slope}=-0.71, \text{intercept}=7.06$).
For the data pooled across all models, this trend continues ($r=-0.26, p<0.0001$).
% , \text{slope}=-2.21, \text{intercept}=16.88$).
Note that although the correlation is statistically significant, the strength is relatively weak, which suggests that the prevalence of bounding boxes in videos cannot be simply predicted by video quality score alone and that other features (e.g. temporal consistency, shape deformation, etc.) may also contribute to the prevalence of bounding box annotations.

\section{How Can We Use GeneVA?}
To demonstrate how GeneVA can be utilized, we introduce a system (Section \ref{sec:artifact_detector}) that automatically detects and describes artifacts in generated videos. These outputs can then, in theory, be used to improve video quality via downstream tasks such as video inpainting.
\subsection{Automated Artifact Detection}
\label{sec:artifact_detector}
\begin{figure}[t]
\centering
\includegraphics[width=\linewidth]{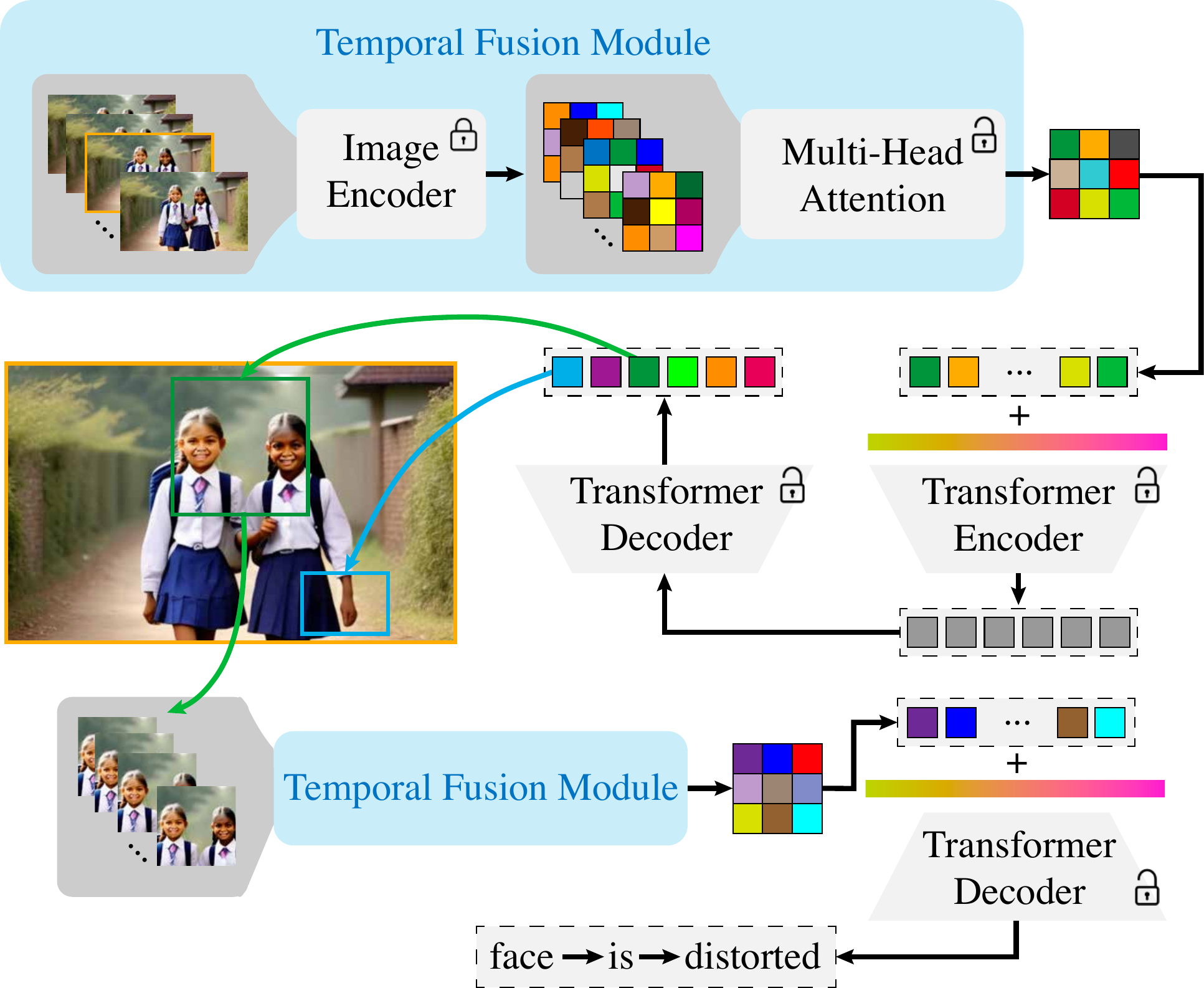}
\caption{
Overview of our proposed pipeline for interpretable artifact detection. First, the model processes a sequence of frames to identify the location of an artifact in the central anchor frame (highlighted in yellow). Following this localization, the system then generates a textual description to explain the nature of the detected artifact.
%Our pipeline for automated artifact detection is illustrated.
}
\label{fig:artifact-detection}
\end{figure}
\begin{figure*}[t]
\centering
\includegraphics[width=\linewidth]{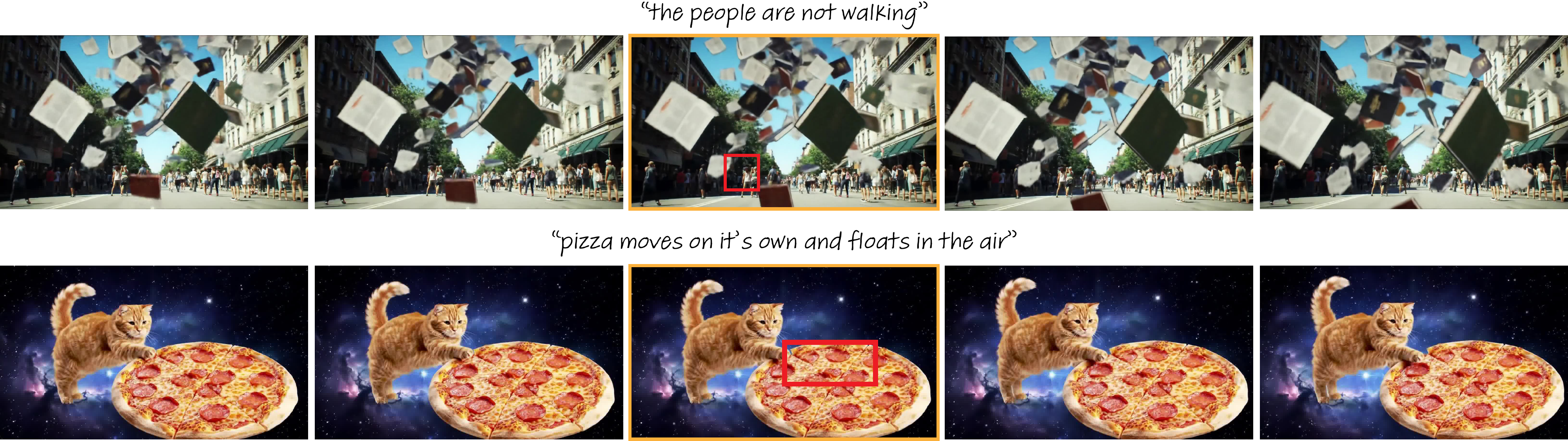}
\caption{
Results of our artifact detection pipeline on an unseen generative model. Our system identifies and describes artifacts by making predictions on the highlighted central frame (in orange) of a video sequence. The detected artifacts are visualized as red bounding boxes, and the predicted descriptions are included above the sequence. The videos are generated by Google's Veo model using text prompts from our dataset.
See the appendix for more results.
}
\label{fig:result-artifact-detection-generalize}
\end{figure*}
Given a video input, our goal is two-fold: 1) identify the location of any immersion-breaking artifact, and 2) describe the type of artifact detected. To accomplish these goals, we propose a dual-stage pipeline that first identifies an artifact's coordinates with the \textbf{Artifact Detector}, then generates a corresponding textual description with the \textbf{Artifact Caption Generator}. To leverage the strength of pre-trained image-based models, we also introduce a \textbf{Temporal Fusion Module} that learns to attend to temporally relevant information across consecutive frames. Figure \ref{fig:artifact-detection} shows the overview of our pipeline.
\subsubsection{Implementation Details}
\paragraph{Temporal Fusion Module}
 Given a video of \(T\) consecutive frames \(\mathcal{V} = \{I_0,\ldots,I_{T-1}\}\), a frozen backbone (e.g. CNN) is used to extract a sequence of per-frame feature vectors, \(\phi(\mathcal{V}) \in \mathbb{R}^{T\times d}\). This is passed through a temporal self-attention layer which projects the features at each spatial location into queries ($Q$), keys ($K$), and values ($V$) and calculates the attention-weighted representation \(\operatorname{Softmax}\!\bigl(QK^{\!\top}/\sqrt{d_h}\bigr)\,V\) that fuses information across frames. Attention parameters are learned jointly during the training of each stage. 
 %Following the corresponding image-based model, we use ResNet-18 and ViT as image encoder for detection and captioning task respectively. 
 In all experiments, we set $T=5$ frames, sampled at 10 FPS, to give the model temporal context for predicting artifacts of the central anchor frame.
\paragraph{Artifact Detector}
%\paragraph{Artifact Detector}
Existing object-detection pipelines often rely on objecness scores (e.g., Faster R-CNN \cite{ren2015faster}, YOLOs \cite{7780460, khanam2024yolov11overviewkeyarchitectural, yolox2021}) which prioritize regions that resemble objects. Video artifacts, however, can emerge anywhere in the frame and may not resemble any conventional objects. To avoid this bias, we adopt the end-to-end transformer-based architecture proposed in RT-DETR \cite{lv2023detrs, lv2024rtdetrv2improvedbaselinebagoffreebies}, which employs an efficient hybrid encoder to achieve state-of-the-art real-time performance. The model is initialized with pre-trained weights (ResNet18 \cite{he2016residual}) and then jointly trained with the temporal fusion module for video artifact localization.

\paragraph{Artifact Caption Generator}
The goal of this stage is to generate a textual description for each artifact region. For training, the model uses regions defined by ground truth annotations, while during inference, it processes the regions provided by our detector. We adopt the GIT architecture \cite{wang2022git}, which utilizes a transformer decoder to autoregressively generate text from visual embeddings. We use the pre-trained weights (Vision Transformer from \cite{wang2022git}) as our initialization and jointly train with our temporal fusion module to generate human-like textual descriptions of the localized artifacts.

\subsubsection{Results}
\begin{table}[tb]
\caption{
Quantitative results for our artifact detection system on the held-out test set. 
We report Average Precision (AP) at specific IoU thresholds (AP$_{25}$, AP$_{50}$, AP$_{75}$). AP denotes the standard COCO metric \cite{coco}, average over IoU thresholds from 0.5 to 0.95 in the step of 0.05. 
%All scores are reported to three decimal places; a value of 0 indicates a result < 0.0005.
}
\resizebox{\linewidth}{!}{
\begin{tabular}{ l | c c c c c c}
\hline
\textbf{Method} & AP$_{.25}$ & AP$_{.50}$ & AP$_{.75}$ & AP  \\ 
\hline
Ours & \textbf{0.132} & \textbf{0.091} & \textbf{0.004} & \textbf{0.032}  \\
w/o temporal fusion & 0.103 & 0.057 & 0  & 0.020 \\
%Random & 0.014 & 0 & 0 & 0  \\
\hline
\end{tabular}
}
\label{table:quan-detect}
\end{table}
For this proof-of-concept study, we randomly sampled a subset of 1,000 videos from our full dataset. This subset is then split at the video level into an 80/10/10 for training, validation, and testing, respectively. We used the validation set to determine the optimal training epoch via early stopping and report final performance on the held-out test set.

As shown in Table \ref{table:quan-detect}, for artifact detection, our system achieves an Average Precision ($AP_{.25}$) of 13\% on the held-out test set. This reflects the model's ability to identify the general location of an artifact, where a detection is considered correct if it overlaps the true region by at least 25\%. This demonstrates that our model is learning a meaningful signal from the data. For context, a random baseline achieves an $AP_{.25}$ of $\approx{1}\%$ on the same task. We also conducted an ablation study using single-frame inputs to isolate the effect of temporal fusion. The lower performance ($-3\%$ in $AP_{.25}$) when removing temporal context indicates that temporal context provides important clues in identifying video artifacts, many of which are inherently temporal in nature.

 While these results are promising, they also highlight the difficulty of the task and the substantial gap that remains to fully match the subtle way humans identify artifacts in generated content. Understanding this aspect is a critical step toward creating a new generation of generative models that align with human perception.

\subsection{Cross-model generalization}
 Our dataset 
 %intentionally designed to be diverse, 
 includes labeled artifacts from three different generative models. The goal is to capture a model-agnostic representation, thereby avoiding overfitting to the unique artifacts of any single generative model. To validate this, we apply our pre-trained artifact detector system (\Cref{sec:artifact_detector}) to videos generated by Google's Veo\footnote{\url{https://deepmind.google/models/veo/}}, a state-of-the-art text to video model \emph{that was not represented in our training data}. Qualitative results are shown in \Cref{fig:result-artifact-detection-generalize}. Our pipeline successfully identifies and provides coherent descriptions for typical artifacts in Veo's outputs, such as unnatural movement -- a common issue across various generative models. This strong zero-shot performance indicates that our dataset encourages the learning of a general representation of video artifacts, making it a robust tool for evaluating novel generative models.
 %We select two prompts from our test set split 
%\paragraph{Quality Metric}
\section{Limitations and Future Work}
% \warning{We don't know the demographics of the users. This is definitely going to create biases in the kinds of videos generated. I (Niall) see this anecdotally from when I was doing my NSFW filtering. There were lots of videos centered on muslim/hindu religion and chinese culture (and lots of sci-fi).}

% it's well-known that identifying artifacts requires a bit of training/expertise. we sort of avoid this problem in this work by having the explicit goal of trying to get a dataset of what the ``average joe'' considers an artifact, but there is likely still some variance within our sampled labelers because people have different levels of familiarity with digital content (eg and older person is probably worse at identifying artifacts than a younger person even if they are the same level of ``average joe'' because the old person does not keep up with technology).

\paragraph{Repeated annotations.}
While conducting annotations under a constrained budget, we faced a key trade-off between the number of unique videos and the number of repeated evaluations per video. For GeneVA, we prioritized maximizing the diversity of videos to capture a broader range of language-conditioned contexts. As a result, the current dataset does not include multiple repeated annotations for the same video. Future extensions incorporating repeated evaluations would enable analysis of inter-observer variability and provide a probabilistic understanding of artifact detectability. Moreover, demographic information was not collected from crowd-sourced participants to maintain privacy. Additional studies with repeated experiments and optional demographic reporting could also extend  data diversity based on randomly sampled participants.

\paragraph{Number of artifacts annotated.}
A key challenge in crowdsourced studies is ensuring data quality through unambiguous task design. In our pilot experiments, we found that participants often had inconsistent interpretations of visual artifacts, particularly given their generative nature. To address this, we instructed users to label no more than five artifacts, encouraging them to focus on the most salient ones, with the order indicating perceived importance. In the future, we plan to develop more adaptive strategies by leveraging insights from the initial GeneVA dataset to predict the likely number of artifacts dynamically.

\paragraph{Potential societal impacts.}
AI-generated text-video content inherently leads to ethical issues, such as the spread of misinformation.
By collecting a dataset of localized artifact labels, future models could potentially be trained to inpaint these artifacts, thereby further accelerating the quality of text-video generation.
We argue that this dataset can be leveraged for good -- better artifact detection, driven by real human annotations, can aid in automatic detection of AI-generated content where it may be difficult to identify.

\section{Conclusion}

We introduced GeneVA, a large-scale, human-annotated dataset to address a critical gap in understanding artifacts in AI-generated video. By combining structured annotations with free-form feedback, GeneVA offers a rich and flexible resource for analyzing and predicting artifact occurrence at multiple levels of detail.
Our proof-of-concept experiments demonstrate its application for training effective artifact prediction models. We anticipate that the public release of GeneVA will accelerate research on evaluating, benchmarking, detecting, and potentially mitigating artifacts in next-generation video generation systems.

{
    \small
    \bibliographystyle{ieeenat_fullname}
    \bibliography{main}
}
\end{document}